# Towards Operational Streamflow Forecasting in the Limpopo River Basin using Long Short-Term Memory Networks

James Tlhomole[1,2], Edoardo Borgomeo[1], Karthikeyan Matheswaran[2],

Mariangel Garcia Andarcia[2]

[1]University of Cambridge

[2]International Water Management Institute

*Corresponding Author*

James Tlhomole: j.tlhomole@gmail.com

## Abstract

Robust hydrological simulation is key for sustainable development, water management strategies, and climate change adaptation. In recent years, deep learning methods have been demonstrated to outperform mechanistic models at the task of hydrological discharge simulation. Adoption of these methods has been catalysed by the proliferation of large sample hydrology datasets, consisting of the observed discharge and meteorological drivers, along with geological and topographical catchment descriptors. Deep learning methods infer rainfall-runoff characteristics that have been shown to generalise across catchments, benefitting from the data diversity in large datasets. Despite this, application to catchments in Africa has been limited. The lack of adoption of deep learning methodologies is primarily due to sparsity or lack of the spatiotemporal observational data required to enable downstream model training. We therefore investigate the application of deep learning models, including LSTMs, for hydrological discharge simulation in the transboundary Limpopo River basin, emphasising application to data scarce regions. We conduct a number of computational experiments primarily focused on assessing the impact of varying the LSTM model input data on performance. Results confirm that data constraints remain the largest obstacle to deep learning applications across African river basins. We further outline the impact of human influence on data-driven modelling which is a commonly overlooked aspect of data-driven large-sample hydrology approaches and investigate solutions for model adaptation under smaller datasets. Additionally, we include recommendations for future efforts towards seasonal hydrological discharge prediction and direct comparison or inclusion of SWAT model outputs, as well as architectural improvements.

# 1. Introduction

Hydrological discharge simulation has become a critical aspect of sustainable water management practices, ensuring equitable use of increasingly stressed water resources under a changing climate (Dembélé et al., 2023). Applications in sustainable water management, risk mitigation and other areas such as hydraulic engineering design closely depend on the quality of hydrological data used. Unfortunately, for many rivers around the world, we only have data from a few points along the river course, typically on the main stem of the river and close to the river mouth, where data have been continuously collected by monitoring stations. Additionally, spatial biases towards large perennial rivers exist globally, with the number of active gauging stations concurrently decreasing since the 1980s (Krabbenhoft et al., 2022, Crochemore et al., 2020). To sustainably manage water resources, however, we need to know the river discharge at different points throughout the river basin, and not just at its main stem, as well as in ungauged basins.

Faced with this issue, hydrologists have typically resorted to simple regression-based estimates of river flows in ungauged tributaries. However, traditional regression-based methods, whilst retaining explainability, often fail to preserve multi-site correlations and do not characterize uncertainty well (Singh et al., 2022). Alternatively, operational physics-based, mechanistic, or process-based models for hydrological discharge prediction have become central to water management at basin, regional and continental scales. These approaches incorporate known physics, employing the conservation of physical quantities, hydrological characteristics, and utilise representations of watershed properties and meteorological conditions as model inputs (Devia et al., 2015). Physics-based models however, are computationally expensive and reliant on catchment-specific calibrations that commonly limit their generalisation beyond calibration sites, or may have unrepresented physical processes that affect model predictive performance (Faramarzi et al., 2015).

Recently, there has been rapid growth in the adoption of data-driven methods for hydrological discharge prediction. Machine learning based methods have been shown to outperform physics-based methods whilst exhibiting greater computational efficiency and improved generalisation in ungauged basins (Kratzert et al., 2019). These methods learn complex nonlinear relationships between input variables and target quantities, directly inferring hydrological processes from large datasets (Razavi et al., 2022). This is as opposed to the explicit encoding of hydrological processes in mechanistic models, which may result in deficient process representation (Faramarzi et al., 2015). Sequential deep learning methods particularly, have been shown to perform well at hydrological discharge prediction. Kratzert et al. (2018) demonstrated the efficacy of the long short-term memory (LSTM) sequential deep learning network architecture at learning rainfall-runoff relationships from data, producing competitive daily forecasts against a SAC-SMA + Snow-17 model. They further showed the ability to uncover learned hydrological behaviour by probing the LSTM internal cell states, improving transparency. Critically, Kratzert et al. (2019) later demonstrated the effectiveness of training models on both meteorological timeseries and catchment attributes, citing enhanced regionalisation due to the ability to learn hydrological similarities and spatial relations. Other deep learning approaches have shown the use of more recent methods such as the incorporation of attention mechanisms and unstructured spatial encoding using graph-based architectures (Yin et al., 2022, Feng et al.).

The adoption of deep learning methodologies has been driven by the proliferation of large sample hydrology datasets such as CAMELS-US (United States) and CAMELS-GB (Great Britain) (Addor et al., 2017, Newman et al., 2015, Coxon et al., 2020). These are posed as datasets of meteorological timeseries, catchment descriptors and observed streamflow. As such, these approaches are commonly limited to data rich regions with the data availability for large dataset curation. This has therefore limited application in data sparse regions, despite the noted outperformance of deep learning methods over statistical and physics-based methods. This disparity is particularly exacerbated within the context of the African continent where few studies have been conducted to investigate the application of these methods. This is despite the urgent need to protection large vulnerable populations from environmental risks, heightening water stress and to inform climate change adaptation. Oruche et al. (2021) investigated the transfer of LSTM models from CAMELS-US to gauge stations in Kenya, investigating the knowledge transfer from large datasets to smaller target domains with sparse observations. Le et al. (2022) similarly investigated prediction in ungauged regions by transferring more traditional machine learning models trained in data rich regions to South America and Southern Africa, emphasising the need to carefully select the data used for pretraining. Different approaches have been employed for the transfer of data-driven models between regions, and it has been posed that training on large heterogeneous data may be beneficial to the data driven methodologies (Ma et al., 2021, Kratzert et al., 2024). More generally, Nearing et al. (2024) presented a global approach for flood prediction utilising an encoder-decoder LSTM architecture and outperforming a GloFAS model using a five-day forecast horizon.

Given the limited application of these methods to data-limited streamflow forecasting in Africa, where timely water management is critical, and the demonstrated outperformance of these methods over traditional regression and physics-based methods, we investigate the application of sequential deep learning methods to the transboundary Limpopo River basin in Southern Africa.

## 2. Methodology

The Limpopo River basin spans an area of approximately 408,250 km$^2$ across four countries; South Africa, Botswana, Zimbabwe and Mozambique. Our approach utilises verified discharge data from Digital Earth Africa (DEA), available at 68 monitoring stations in the South African side of the basin at daily timescale, as shown in Figure 1a. The recording period of the observed discharge data spans a period of approximately 23.8 years (01/01/2001 – 30/10/2024), with an estimate of roughly 18% missing data. From this we select a subset of 53 stations based on data availability and the supplied data quality codes. Data partitioning is based on previous studies where models are typically trained and evaluated on approximately 10 years of continuous data (Train: 01/01/2001 - 30/09/2010, Validation: 01/10/2010 - 30/09/2013, Test: 01/10/2013 - 30/10/2024).

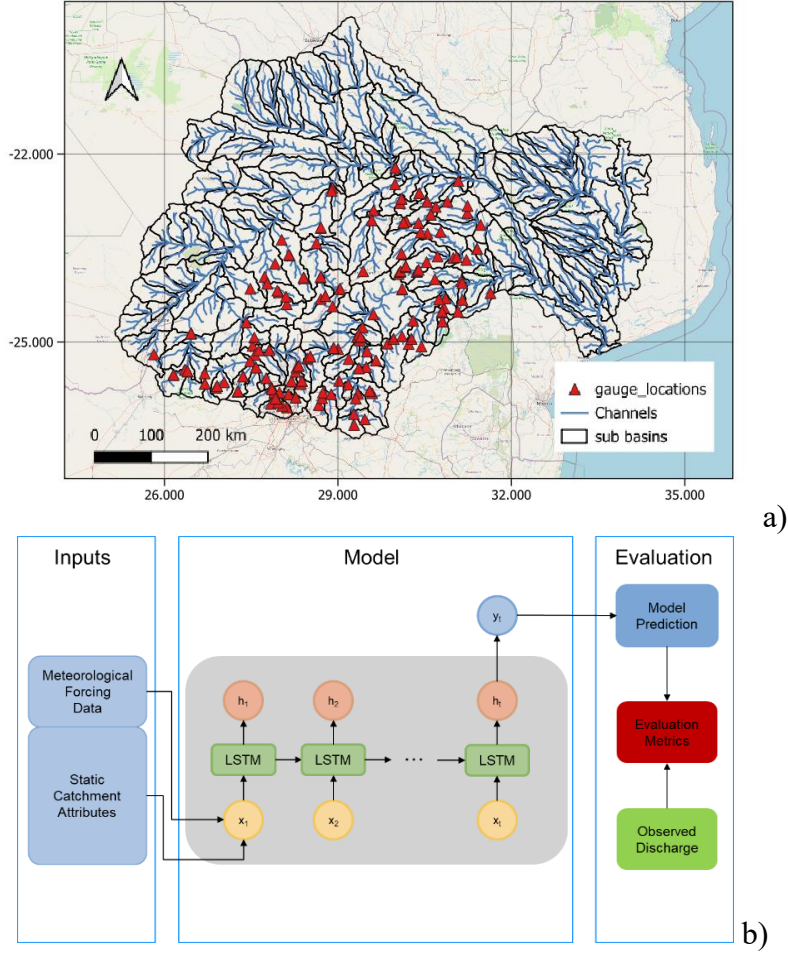

*Figure 1: a) Spatial map of verified discharge gauges in the Limpopo river basin, b) General workflow adopted.*

## 2.1. Computational experiments

Since this investigation is centred around the application of current state of the art approaches to data limited regions in Africa, we investigate the use of dominant deep learning approaches at hydrological discharge forecasting in the Limpopo River basin. The adopted model architecture is based on a single layer LSTM (Long Short-Term Memory) approach, which has been shown to outperform traditional methods across large-sample hydrology studies. We utilise a single layer LSTM with a hidden size of 256. Computational experiments conducted in this study involve augmenting the input feature space between experiments by varying the meteorological, static and dam release data used for model training as outlined below. Models process historical timeseries with a sequence length 30 days for each feature investigated as input. Experiments are posed as sequence to value problems utilising an LSTM to map from a sequence of antecedent conditions to a single output discharge value at the next timestep, as visualised in Figure 1b. The choice of 30-day input sequence length, 256 hidden size and number of LSTM layers represent hyperparameters that can further be tuned. Model evaluation is carried out by evaluating the Nash-Sutcliffe Efficiency (NSE) of the predicted and observed discharge, as shown in Figure 1b. We outline the computational experiments undertaken below.

**All stations (Basin-wide approaches):**

For these investigations, models are trained on data from all stations simultaneously.

- **CHIRPS Precipitation:** Initial experiments involved training the LSTM models on pointwise timeseries of daily CHIRPS precipitation. This is the same product utilised in the Limpopo operational SWAT+ model and provides a rainfall-runoff baseline i.e. represents the model's ability to learn a direct general nonlinear mapping from precipitation to runoff. For each station, the model processes a timeseries of previous precipitation extracted for the given station and predicts the observed discharge at the next timestep.

- **CHIRPS + ERA5 Dynamic:** This experiment involves an extension of the above with the addition of variables from ERA5-Land. This includes 2m temperature, surface net solar radiation, surface net thermal radiation, total precipitation, and surface pressure. This subset of variables is determined from an approach shown to perform well globally (Nearing et al. 2024). The use of multiple precipitation products i.e. CHIRPS and ERA5-Land, has also been shown to produce potential synergy (Kratzert et al., 2021).

- **CARAVAN:** This approach utilises open-source software for extending the CARAVAN dataset with additional observed streamflow data at new gauge stations (Kratzert et al., 2023). This standardises the extraction and aggregation of meteorological drivers and static catchment attributes for new sites and facilitates potential integration into global modelling approaches. Similar to the above, meteorological variables are derived from ERA5-Land, using the same subset from the global approach of Nearing et al., (2024); 2m temperature, surface net solar radiation, surface net thermal radiation, total precipitation, and surface pressure. Additionally, we utilise the set of static catchment attributes reported to have the highest feature importance from the global approach, with the exception of Snow Cover Extent, Permafrost Extent and Glacier Extent.

**Station Specific Experiments:**

These investigations focus on specific gauges.

- **Dam Release:** The Limpopo river basin is significantly regulated, with at least 345 dams in the basin. Typically, data-driven approaches that have been shown to outperform mechanistic models in literature are trained and evaluated on naturalised river flows and are not applied where there are significant abstractions, dams, and human impacts. These disturbances to the natural river flows may hamper the ability of models to learn the rainfall-runoff relationships directly from the data. Therefore, we investigate the effect of including remote-sensing-derived dam release information as an additional input to the data-driven model as a proxy for the divergence from naturalised flow conditions. For this, we include the "dam level below max" available at daily resolution from the digital twin data API, into the data driven model input space.

The model is the same baseline as the basin-wide CHIRPS + ERA5 Dynamic experiment above, using the same aforementioned input variables, with the addition of the dam release proxy data. We apply this approach to data derived for a single dam; Haartbeespoort Dam, and observed discharge from station 146 downstream of the dam, as shown in Figure 4.

- **Hybrid:** Rather than utilising a purely data-driven method, as is the case with the other computational experiments, this approach attempts to combine the outputs of the physics-based SWAT+ model that is currently deployed in the digital twin platform, with the data-driven method, underpinning the hybrid approach. The LSTM model in this case can be thought of as a data-driven model correction to the SWAT model output, thus retaining rather than discarding the physics of the SWAT model approach. The SWAT natural flow also provides a proxy for the naturalised flows in the basin, which should allow the LSTM to learn abstraction effects and thus the modified rainfall-runoff relationships. Here we focus on forecasting for a single gauge station 272 (A7H008), SWAT channel 215, although the method can be scaled towards a basin-wide approach by considering all channels. A single LSTM model is trained to predict the month-ahead average discharge at the gauge station. The model takes, as input, timeseries of 1) monthly cumulative CHIRPS precipitation and 2) SWAT natural model output from the data API, for the preceding 12-month period, and outputs the average discharge for next month. As such, this model performs forecasting at monthly timesteps rather than the daily timesteps of all previous approaches.

## 3. Results

### 3.1 All Stations (Basin-Wide approaches)

Table 1: Summary of model performance for all stations (basin wide approach) over the test period.

| Model | NSE | | No. of basins NSE < 0 |
|---|---|---|---|
| | mean | median | |
| CHIRPS | -4.688 | -0.378 | 41 |
| CHIRPS + ERA5 | -4.115 | -0.694 | 36 |
| CARAVAN | -0.116 | 0.109 | 16 |

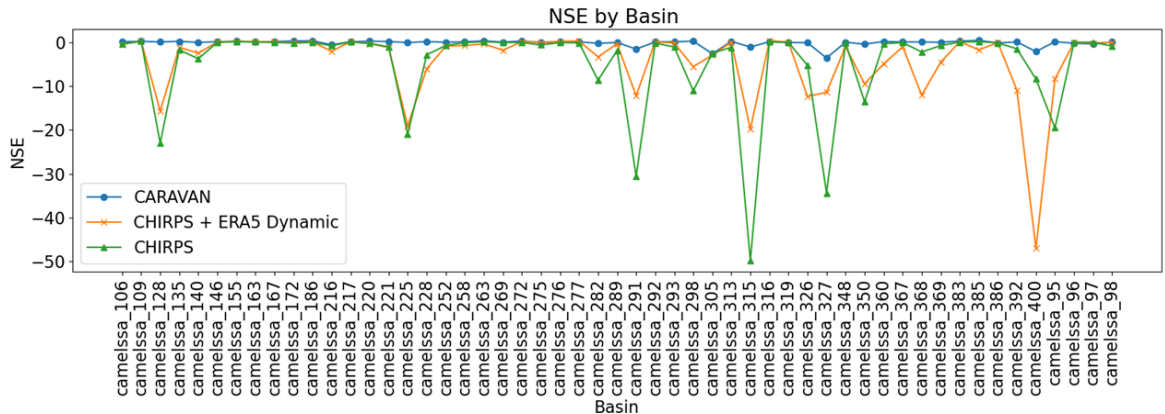

*Figure 2: Plot of NSE values for all stations (basin wide approach) evaluated at each station for the test period and for three computational experiments using different input data: CHIRPS precipitation only, CHIRPS and dynamic drivers derived from ERA5-Land, and the input feature set from CARAVAN; ERA5-Land dynamic drivers and static catchment descriptors.*

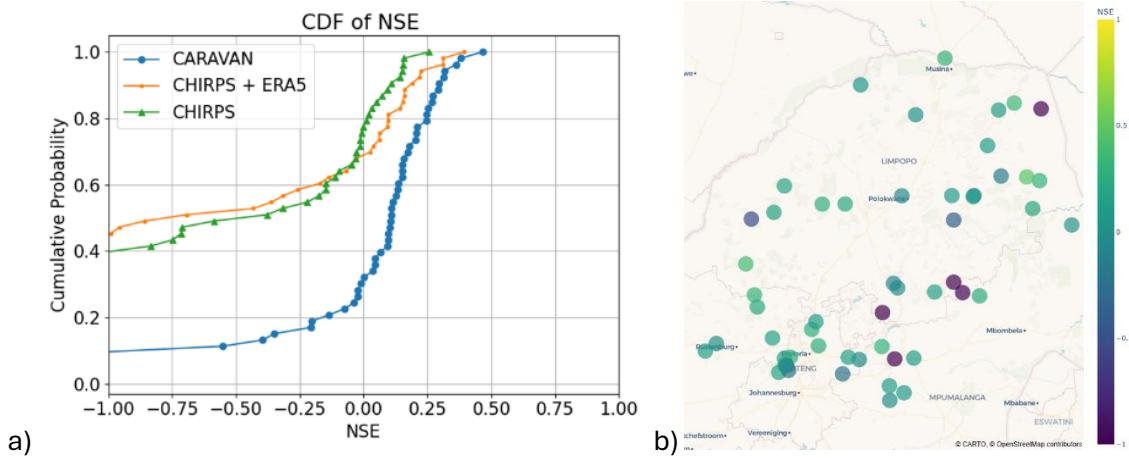

*Figure 3: Model performance for the test period evaluation. a) Cumulative density function (CDF) of model performance for all station (basin wide) approaches, and b) spatial distribution of NSE values for the model trained with the CARAVAN approach.*

Figure 2 shows the performance of the basin-scale models trained using the different input data. The modelling approach remains unchanged while the input data to the model is varied between experiments. The results show the performance at each station of the models trained across all stations using different input feature sets; 1) CHIRPS precipitation only, 2) CHIRPS precipitation and dynamic meteorological drivers extracted from ERA5-Land, and 3) the input feature set from CARAVAN; ERA5-Land meteorological drivers and static catchment descriptors.

The CARAVAN methodology consistently outperforms the models without the addition of static attributes. The model trained with CHIRPS and ERA5 Dynamic variables typically performs better than the model trained with only CHIRPS precipitation, suggesting benefits to having multiple precipitation products as well as further meteorological descriptors. The relative performance of the basin wide-models appears to suffer at the same stations suggesting systematic difficulties in the application of data-driven models at those specific stations, indicating the presence of processes that are not captured sufficiently by meteorological and static descriptors only. The outperformance of CARAVAN over the other approaches is also shown in Table 1, where it was the only approach that yielded a positive median NSE value and the lowest number of basins with negative NSE values. Figure 3a shows the performance of the different models as a cumulative density function, showing that the CARAVAN approach was able to achieve a maximum NSE of around 0.48, while the combined CHIRPS and ERA5 dynamic approach achieved 0.36 and the CHIRPS only approach around 0.25.

## 3.2 Station Specific Experiments

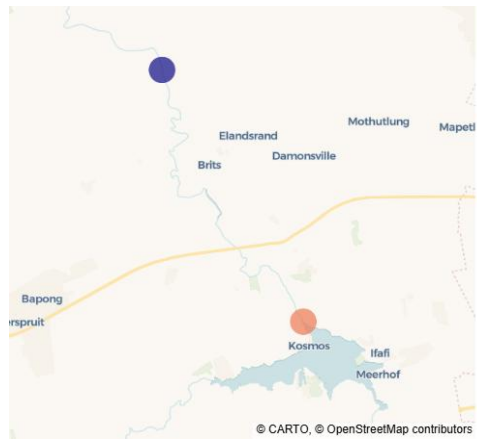

Figure 4: Example of reservoir impacts in the Limpopo River basin. The observed discharge gauge station (dark purple) of interest is directly downstream of Haartbeespoort Dam (orange).

Table 2: Summary of model performance for the station directly downstream of Haartbeespoort Dam.

| Model | NSE (Basin Wide) | NSE (Single Station) |
| --- | --- | --- |
| CHIRPS + ERA5 | 0.006 | 0.313 |
| CHIRPS + ERA5 + Dam Release | - | 0.416 |
| CARAVAN | 0.264 | 0.383 |

Table 2 shows the effect on model performance of the addition of remote-sensing-derived dam release data, for Haartbeespoort dam, at a gauge station downstream of the dam. It is evident that the addition of dam release data as a proxy for commonly overlooked human impacts in data-driven approaches leads to an NSE improvement of 0.103. For this case, training on the individual station downstream of the dam, rather than for the entire basin, leads to an improvement across all approaches. This suggests that the models may sacrifice performance at certain individual stations in order to improve overall basin wide performance, which may be particularly troublesome where dam releases are involved.

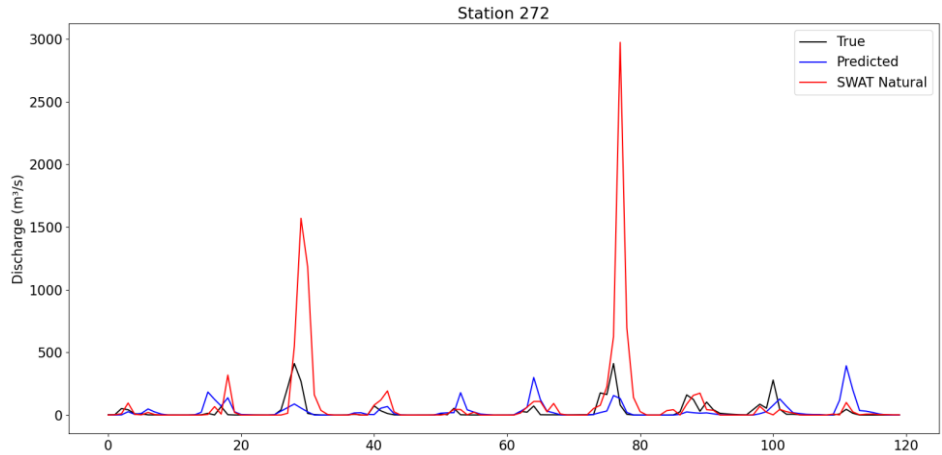

*Figure 5: Hybrid approach evaluated at gauge station 272 (A7H008), SWAT channel 215.*

Figure 5 shows the performance of the model trained to predict the average discharge of the month ahead, based on an input of monthly cumulative CHIRPS precipitation and the SWAT natural model output from the digital twin data API, for the preceding 12-month period. The model is trained at gauge station 272 (A7H008), SWAT channel 215. The LSTM model is posed as a data-driven correction to the SWAT model output, and the SWAT model in this hybrid approach can be seen to provide a proxy for naturalised flows. Although this approach has been included as an initial concept for the hybrid approach, it serves as a pathway towards seasonal prediction in the Limpopo river basin. As such, the current model may struggle to predict the discharge at a monthly timestep as opposed to the daily timestep of the previous approaches, particularly since this does not include the forecasted meteorological conditions over the basin for the longer prediction horizon. Figure 6 shows the proposed improved model architecture for this purpose. Specifically, the proposed model adaptations for transitioning to seasonal prediction include using the CARAVAN framework as an encoder to process historical meteorological or hindcast data as well as static descriptors. This then passes the hidden state of the encoder to a decoder model that processes the forecasted meteorological drivers. The latter may be similar to feeding seasonal precipitation forecasts to the SWAT model to produce seasonal discharge predictions. This may also provide a flexible way to process the SWAT model outputs in a hybrid approach by parsing the SWAT natural forecasts along with the forecast meteorological data. Such a method may indeed be of greater interest for water management in the Limpopo river basin than the previous daily discharge estimation methods.

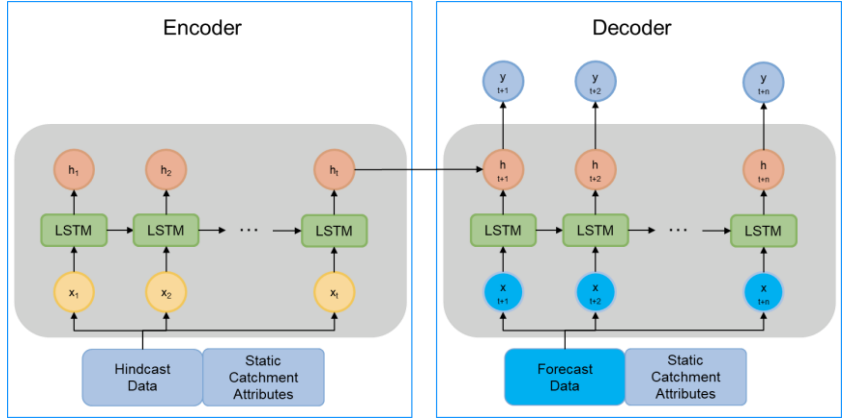

*Figure 6: Proposed architecture for improved hybrid approach; towards seasonal forecasting in the Limpopo River basin at monthly timesteps.*

## 4. Discussion

The work undertaken in this investigation largely explores how data driven methods, particularly LSTMs, that have been shown to outperform mechanistic or physics-based methods at streamflow forecasting, perform when applied to the Limpopo River basin. The Limpopo River basin presents a challenging application site characterised by significant human impacts from irrigation, the arid or semi-arid nature of the basin, and the presence of dams which are commonly neglected in large sample hydrology studies (Addor et al., 2020, Tran et al., 2025). Our results show that these purely data-driven approaches tend to struggle in the Limpopo due to these factors, despite outperformance on global models reported in the literature (e.g., Nearing). This is contrary to the literature that shows that these methods benefit from training on larger more diverse datasets and perhaps indicates the need to carefully select the basins that form the pretraining dataset, either through spectral similarity methods or similarity of the hydrological regimes. The choice of precipitation product may be a significant source of uncertainty and merits further investigation, particularly over the arid/semi-arid regions of Southern Africa where rainfall will be driven by convective processes, and where there may be insufficient in-situ measurements to constrain models and reanalysis products. The use of ERA5-Land within the global CARAVAN approach has been selected due to its global availability but may further hamper the performance of a data driven methodology over the Limpopo river basin. This will also become increasingly important when transitioning to seasonal predictions due to the growing uncertainty with precipitation forecast lead time, translating directly into streamflow uncertainty. In this case however, data driven methods may provide fast means for generating ensembles of streamflow predictions, aiding water management. The river basin also exhibits a large number of discharge gauge stations with long periods of zero flows, where the data driven model will struggle. In addition to the presence of a number of reservoirs on the South African side of the basin which control the flow, large inter-basin transfers and other dynamics may also represent significant process within the Limpopo that the models do not adequately capture from the available model inputs, limiting the model reliability.

The inclusion of dam release data into the model input space for prediction downstream of Haartbeespoort dam was seen to produce improvement in the model prediction NSE. A suggested improvement could be inclusion of remote-sensing-derived dam releases at the basin-scale, which could be achieved through a separate "Dam encoder" module that would parse the dam level timeseries for all of the dams simultaneously. This is as opposed to adding individual dam release data directly to the input space of the LSTM for a single downstream gauge, which was investigated here. Instead, the encoder module would parse the timeseries of the dam release, as well as the relevant dam metadata (location, distance, capacity, binary flag for upstream/downstream), before concatenating with the output of the current LSTM which attempts to learn the rainfall-runoff response as was investigated in this study. Such an approach can also be used to predict the basin-wide streamflow discharges simultaneously, akin to how the SWAT model predicts all channels simultaneously, rather than at individual stations as is currently done in this study. Similarly, graph neural network (GNN) architectures may also be advantageous to this study site due to the unstructured nature of the distribution of gauges throughout the basin. Since the architecture has shown dominance at message passing, this could also be beneficial for learning the routing of water volumes between different parts of the basin. Additionally, the computational graph can also be designed to include distinctive nodes that differentiate between dams and gauges, providing a way to inherently include the dam release data and volume transfer at basin scale. This may be better suited to learning the routing and volume transfer between the different gauge stations.

As mentioned in the Results section, transitioning the current method from day-ahead prediction to seasonal forecasting would involve the addition of a decoder that parses the seasonal meteorological forecast. This is similar to the approach of Nearing et al., (2024); the primary difference being that the cited study used 7 days of forecast data at a daily timestep, whereas a monthly timestep and seasonal forecast would be utilised for prediction at water management timescales of interest. Primarily the challenge with this investigation has been the limited data availability, both temporally and spatially. Additionally, the available discharge observations are sourced entirely from the South African side of the transboundary river basin. Such approaches may improve with the construction of African large sample hydrology datasets, with frameworks such as CARAVAN providing a flexible method of adding new data as it becomes available, and pave the way towards democratising streamflow forecasting in regions without the computational power or expertise to operationalise and calibrate physics-based approaches.

## 5. Conclusions

The Limpopo River basin is a vastly important site due to its transboundary nature, the large population it supports and its influence across a range of activities including agriculture. This investigation explores the application of data-driven methodologies, that have been shown to outperform physics-based methods, to the data-limited African river basin. Our results showed that these methods struggled to capture the significant dynamics that contribute to the streamflow in the river basin, perhaps including inter-basin transfers. Future work will include seasonal prediction for water management, alternative deep learning architectures and the development of proxies for significant hydrological processes that are unaccounted for in the current model implementation.